\documentclass[10pt,twocolumn,letterpaper]{article}

\usepackage{iccv}
\usepackage{times}
\usepackage{graphicx}
\usepackage{amsmath}
\usepackage{amssymb}
\usepackage{dsfont}
\usepackage{algorithm}
\usepackage{algorithmic}
\usepackage{multirow}
\usepackage{booktabs}
\usepackage{adjustbox}
\usepackage{subfigure}

\usepackage[dvipsnames]{xcolor}
\usepackage[accsupp]{axessibility}

\usepackage[pagebackref=true,breaklinks=true,letterpaper=true,colorlinks,bookmarks=false]{hyperref}

\iccvfinalcopy 

\def\iccvPaperID{8581} 

\ificcvfinal\fi

\begin{document}

\title{Self-supervised Product Quantization for Deep Unsupervised Image Retrieval}

\author{Young Kyun Jang \and Nam Ik Cho \and\\
Department of ECE, INMC, Seoul National University, Seoul Korea\\
{\tt\small kyun0914@ispl.snu.ac.kr, nicho@snu.ac.kr}
}
\maketitle
\ificcvfinal\thispagestyle{empty}\fi

\begin{abstract}
Supervised deep learning-based hash and vector quantization are enabling fast and large-scale image retrieval systems. By fully exploiting label annotations, they are achieving outstanding retrieval performances compared to the conventional methods. However, it is painstaking to assign labels precisely for a vast amount of training data, and also, the annotation process is error-prone. To tackle these issues, we propose the first deep unsupervised image retrieval method dubbed $\textbf{S}$elf-supervised $\textbf{P}$roduct $\textbf{Q}$uantization (SPQ) network, which is label-free and trained in a self-supervised manner. We design a Cross Quantized Contrastive learning strategy that jointly learns codewords and deep visual descriptors by comparing individually transformed images (views). Our method analyzes the image contents to extract descriptive features, allowing us to understand image representations for accurate retrieval. By conducting extensive experiments on benchmarks, we demonstrate that the proposed method yields state-of-the-art results even without supervised pretraining.
\end{abstract}

\section{Introduction}

Approximate Nearest Neighbor (ANN) search has received much attention in image retrieval research due to its low storage cost and fast search speed. There are two mainstream approaches in the ANN research, one is \textit{Hashing}~\cite{Survey}, and the other is \textit{Vector Quantization} (VQ)~\cite{VQ}. Both methods aim to transform high-dimensional image data into compact binary codes while preserving the semantic similarity, where the difference lies in measuring the distance between the binary codes.

\begin{figure}[!t]
\centering
\subfigure[Contrastive Learning]{
\includegraphics[width=0.9\linewidth]{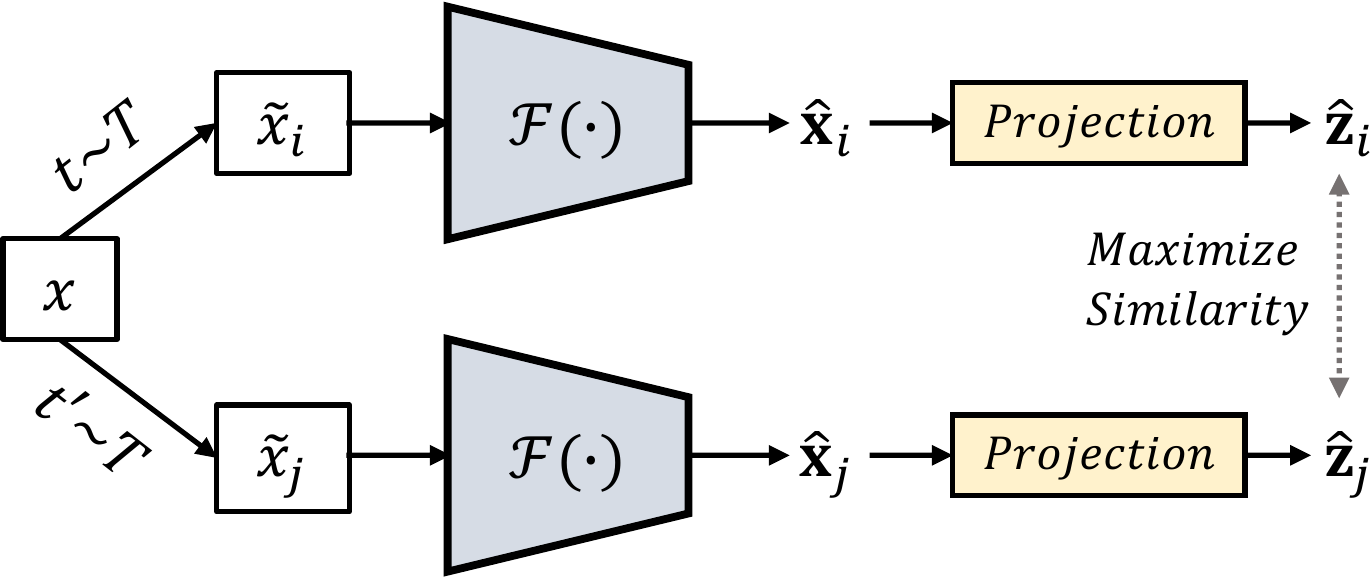}
\label{fig:Figure1_a}
}
\subfigure[Cross Quantized Contrastive Learning (Ours)]{
\includegraphics[width=0.9\linewidth]{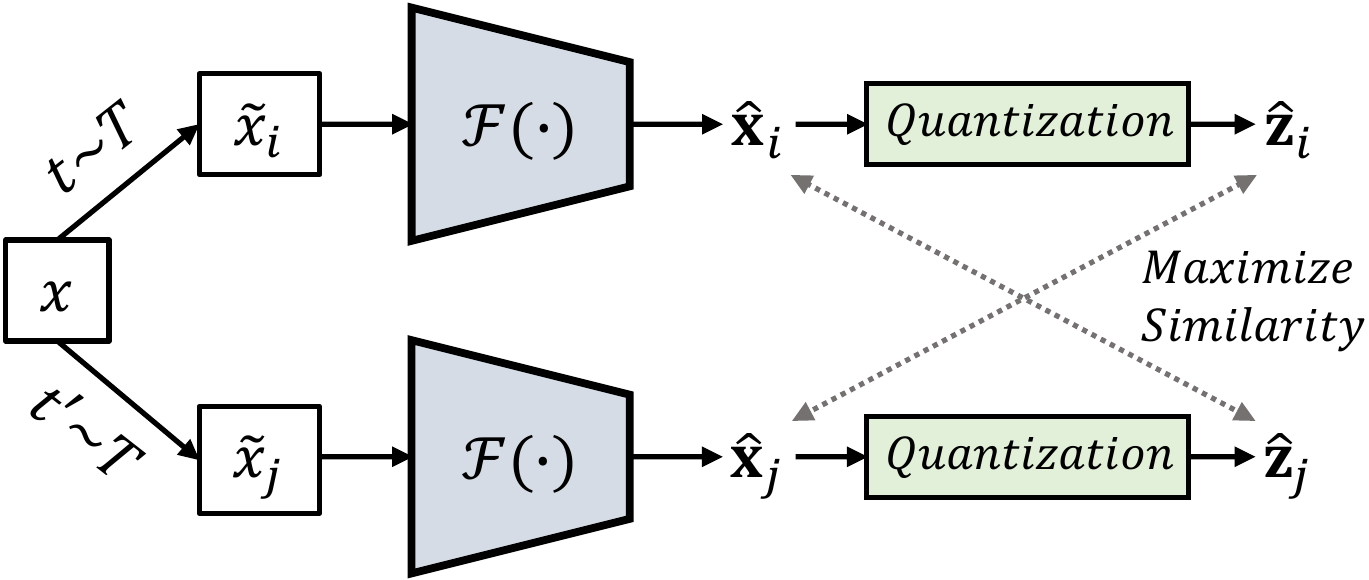}
\label{fig:Figure1_b}
}
\caption{Comparison between (a) contrastive learning and (b) cross quantized contrastive learning. The separately sampled two transformations ($t, t'\sim\mathcal{T}$) are applied on an image $x$ to generate two different views $\tilde{x}_i$ and $\tilde{x}_j$, and corresponding deep descriptor $\mathbf{\hat{x}}_i$ and $\mathbf{\hat{x}}_j$ are obtained from the feature extractor $\mathcal{F}(\cdot)$, respectively. The feature representations in contrastive learning are achieved by comparing the similarity between the projection head outputs $\mathbf{\hat{z}}_i$ and $\mathbf{\hat{z}}_j$. Instead of projection, we introduce the quantization head, which collects codebooks of product quantization. By maximizing cross-similarity between the deep descriptor of one view and the product quantized descriptor of the other, both codewords and deep descriptors are jointly trained to contain discriminative image content representations.} 
\label{fig:Figure1}
\end{figure}

In the case of hashing methods~\cite{LSH, SH, KMH, ITQ, DGH}, the distance between binary codes is calculated using the Hamming distance, {\em i.e.,} a simple XOR operation. However, this approach has a limitation that the distance can be represented with only a few distinct values, where the complex distance representation is incapable. To alleviate this issue, VQ-based methods \cite{PQ, OPQ, LOPQ, AQ, CQ, TQ, SQ} have been proposed, exploiting quantized real-valued vectors in distance measurement instead. Among these, \textit{Product Quantization} (PQ)~\cite{PQ} is one of the best methods, delivering the retrieval results very fast and accurately.

The essence of PQ is to decompose a high-dimensional space of feature vectors (image descriptors) into a Cartesian product of several subspaces. Then, each of the image descriptors is divided into several subvectors according to the subspaces, and the subvectors are clustered to form centroids. As a result, \textit{Codebook} of each subspace is configured with corresponding centroids (\textit{codewords}), which are regarded as quantized representations of the images. The distance between two different binary codes in the PQ scheme is asymmetrically approximated by utilizing real-valued codewords with look-up table, resulting in richer distance representations than the hashing. 

Recently, supervised deep hashing methods \cite{CNNH, HashNet, DCBH, DSDH, CSQ} show promising results for large-scale image retrieval systems.  However, since binary hash codes cannot be directly applied to learn deep continuous representations, performance degradation is inevitable compared to the retrieval using real vectors. To address this problem, quantization-based deep image retrieval approaches have been proposed in \cite{SUBIC, DQN, PQN, DTQ, DPQ, GPQ}. By introducing differentiable quantization methods on continuous deep image feature vectors (deep descriptors), direct learning of deep representations is allowed in the real-valued space.

Although deep supervised image retrieval systems provide outstanding performances, they need expensive training data annotations.
Hence, deep unsupervised hashing methods have also been proposed~\cite{DeepBit, SGH, BGAN, BinGAN, GreedyHash, HashGAN, DVB, DistillHash, TBH}, which investigate the image similarity to discover semantically distinguishable binary codes without annotations. 
However, while quantization-based methods have advantages over hashing-based ones, only limited studies exist that adopt quantization for deep unsupervised retrieval. For example, \cite{UNQ} employed pre-extracted visual descriptors instead of images for the unsupervised quantization.

In this paper, we propose the first \textit{unsupervised} end-to-end deep quantization-based image retrieval method; \textit{Self-supervised Product Quantization} (SPQ) network, which jointly learns the feature extractor and the codewords. As shown in Figure \ref{fig:Figure1}, the main idea of SPQ is based on self-supervised contrastive learning \cite{Simclr, CURL, SWAV}. We regard that two different ``views'' (individually transformed outputs) of a single image are correlated, and conversely, the views generated from other images are uncorrelated. To train PQ codewords, we introduce \textit{Cross Quantized Contrastive learning}, which maximizes the cross-similarity between the correlated deep descriptor and the product quantized descriptor. This strategy leads both deep descriptors and PQ codewords to become discriminative, allowing the SPQ framework to achieve high retrieval accuracy.

To demonstrate the efficiency of our proposal, we conduct experiments under various training conditions. Specifically, unlike previous methods that utilize pretrained model weights learned from a large labeled dataset, we conduct experiments with ``truly'' unsupervised settings where human supervision is excluded. Despite the absence of label information, SPQ achieves state-of-the-art performance.

The contributions of our work are summarized as:

\begin{itemize}
\item To the best of our knowledge, SPQ is the first deep unsupervised quantization-based image retrieval scheme, where both feature extraction and quantization are included in a single framework and trained in a self-supervised fashion.
\item By introducing cross quantized contrastive learning strategy, the deep descriptors and the PQ codewords are jointly learned from two different views, delivering discriminative representations to obtain high retrieval scores.
\item Extensive experiments on fast image retrieval protocol datasets verify that our SPQ shows state-of-the-art retrieval performance even for the truly unsupervised settings.
\end{itemize}

\section{Related Works}

This section categorizes image retrieval algorithms regarding whether or not deep learning is utilized (conventional versus deep methods) and briefly explains the approaches. For a more comprehensive understanding, refer to a survey paper~\cite{Survey}.

\noindent\textbf{Conventional methods.} One of the most common strategies for fast image retrieval is hashing. For some examples, Locality Sensitivity Hashing (LSH)~\cite{LSH} employed random linear projections to hash. Spectral Hashing (SH) \cite{SH} and Discrete Graph Hashing (DGH) \cite{DGH} exploited graph-based approaches to preserve data similarity of the original feature space. K-means Hashing (KMH) \cite{KMH} and Iterative Quantization (ITQ) \cite{ITQ} focused on minimizing quantization errors that occur when mapping the original feature to discrete binary codes.

Another fast image retrieval strategy is vector quantization. There are Product Quantization (PQ) \cite{PQ} and its improved variants; Optimized PQ (OPQ) \cite{OPQ}, Locally Optimized PQ (LOPQ) \cite{LOPQ}, and methods with different quantizers, such as Additive \cite{AQ}, Composite \cite{CQ}, Tree \cite{TQ}, and Sparse Composite Quantizers \cite{SQ}. Our SPQ belongs to the PQ family, where the deep feature data space is divided into several disjoint subspaces. The divided deep subvectors are then trained with our proposed loss function to find the optimal codewords.

\begin{figure*}[!t]
\centering
\includegraphics[width=0.99\linewidth]{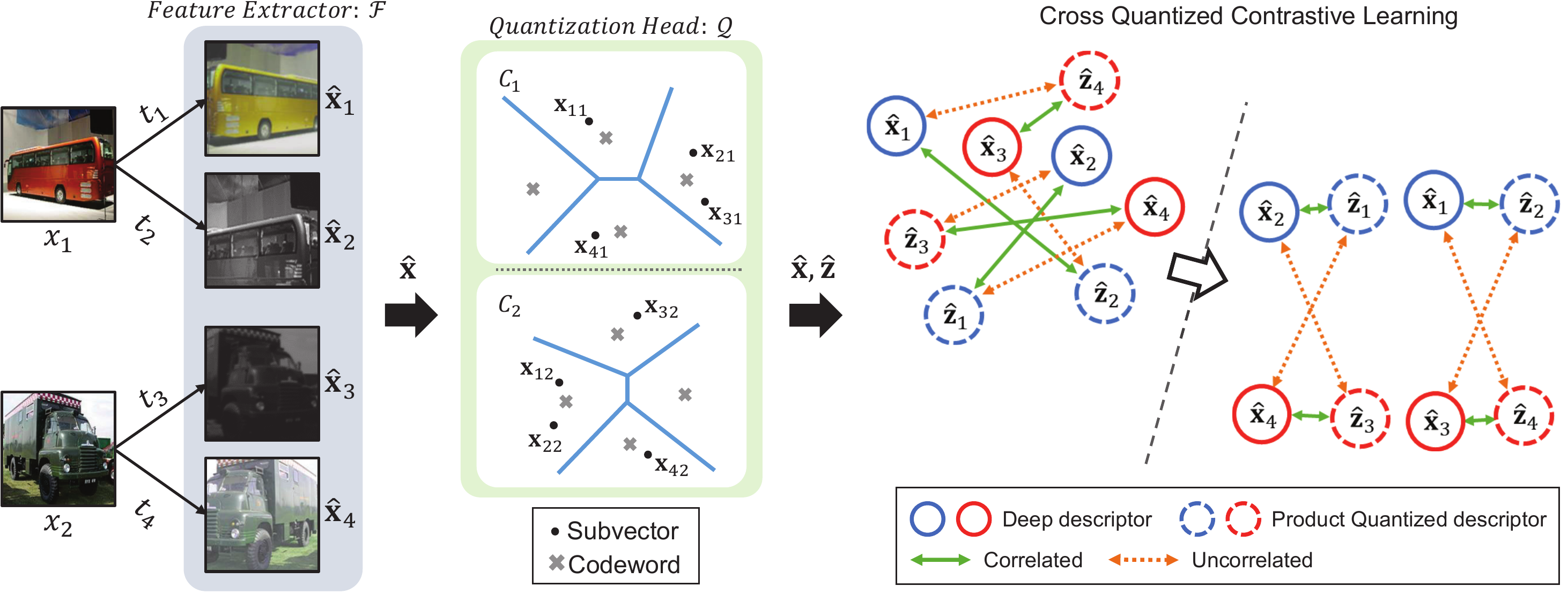}
\caption{An illustration of feature extraction, quantization, and training procedure in SPQ. Randomly sampled data augmentation techniques $(t_n \sim T)$ are applied on $x_1$ and $x_2$ to produce transformed images (different views). There are two trainable components; (1) CNN-based feature extractor $\mathcal{F}$, and (2) quantization head $\mathcal{Q}$, which collects multiple codebooks to conduct product quantization. For example, we set up two codebooks $C_1$ and $C_2$, and illustrate 2D conceptual Voronoi diagram in $\mathcal{Q}$. The original feature space of deep descriptor (feature vector $\mathbf{\hat{x}}_n\in R^D$) is divided into two subspaces and generates subvectors; $\mathbf{x}_{nm}$ where $m=\{1,2\}$ and $\mathbf{x}_{nm}\in R^{D/2}$. By employing soft quantizer $q_m(\cdot)$ on each $\mathbf{x}_{nm}$, the sub-quantized descriptor $\mathbf{z}_{nm}=q_m(\mathbf{x}_{nm})$ is approximated with the combination of the codewords. Notably, subvectors representing similar features are allocated to the same codeword. The output product quantized descriptor $\mathbf{\hat{z}}_n\in R^D$ is obtained by concatenating the sub-quantized descriptors along the $D$-dimension. For better understanding, we paint the feature representations related to $x_1$ in \color{RoyalBlue}blue\color{black}, and $x_2$ in \color{red}red\color{black}. Taking into account the cross-similarity between $\mathbf{\hat{x}}$ and $\mathbf{\hat{z}}$ as: $\mathbf{\hat{x}_1}\leftrightarrow\{\mathbf{\hat{z}_2},\mathbf{\hat{z}_4}\}$,
$\mathbf{\hat{x}_2}\leftrightarrow\{\mathbf{\hat{z}_1},\mathbf{\hat{z}_3}\}$,
$\mathbf{\hat{x}_3}\leftrightarrow\{\mathbf{\hat{z}_2},\mathbf{\hat{z}_4}\}$, and $\mathbf{\hat{x}_4}\leftrightarrow\{\mathbf{\hat{z}_1},\mathbf{\hat{z}_3}\}$, the network is trained to understand the discriminative image contents, while simultaneously collecting frequently occurring local patterns into the codewords.}
\label{fig:Figure2}
\end{figure*}

\noindent\textbf{Deep methods.} Supervised deep convolutional neural network (CNN)-based hashing approaches \cite{CNNH, HashNet, DCBH, DSDH, CSQ} have shown superior performance in many image retrieval tasks. There are also quantization-based deep image retrieval methods \cite{DQN, DPQ}, which use pretrained CNNs and fine-tune the network to train robust codewords together. For improvement, the metric learning schemes are applied in \cite{PQN, DTQ, GPQ} to learn codewords and deep representations together with the pairwise semantic similarity. Note that we also utilize a type of metric learning, {\em i.e.,} contrastive learning; however, our method requires no label information in learning the codewords.

Regarding unsupervised deep image retrieval, most works are based on hashing. To be specific, generative mechanisms are utilized in \cite{SGH, BGAN, BinGAN, HashGAN}, and graph-based techniques are employed in \cite{DVB, TBH}. Notably, DeepBit \cite{DeepBit} has a similar concept to SPQ in that the distance between the transformed image and the original one is minimized. However, the hash code representation has a limitation in that only a simple rotational transformation is exploited. In terms of deep quantization, there only exists a study dubbed Unsupervised Neural Quantization (UNQ) \cite{UNQ}, which uses pre-extracted visual descriptors instead of employing the image itself to find the codewords.

To improve the quality of image descriptors and codewords for unsupervised deep PQ-based retrieval, we configure SPQ with a feature extractor to explore the entire image information. Then, we jointly learn every component of SPQ in a self-supervised fashion. Similar to \cite{Simclr, CURL, SWAV}, the full knowledge of the dataset is augmented with several transformations such as crop and resize, flip, color distortion, and Gaussian blurring. By cross-contrasting differently augmented images, both image descriptors and codewords become discriminative to achieve a high retrieval score.


\section{Self-supervised Product Quantization}

\subsection{Overall Framework}
\label{subsection3.1}

The goal of an image retrieval model is to learn a mapping $\mathcal{R}:x \mapsto \mathbf{b}$ where $\mathcal{R}$ denotes the overall system, $x$ is an image included in a dataset $\mathcal{X}=\{x_n\}^N_{n=1}$ of $N$ training samples, and $\mathbf{\hat{b}}$ is a $\mathrm{B}$-bits binary code $\mathbf{\hat{b}}\in\{0,1\}^\mathrm{B}$. As shown in Figure \ref{fig:Figure2}, $\mathcal{R}$ of the SPQ contains a deep CNN-based feature extractor $\mathcal{F}(x;\theta_{\mathcal{F}})$ which outputs a compact deep descriptor (feature vector) $\mathbf{x}\in R^D$. Any CNN architecture can be exploited as a feature extractor, as long as it can handle fully connected layer, e.g. AlexNet \cite{AlexNet}, VGG \cite{VGG}, or ResNet \cite{ResNet}. We configure the baseline network architecture with ResNet50 that generally shows outstanding performance in image representation learning, and details are reported in section \ref{subsection4.2}.

Regarding the quantization for fast image retrieval, $\mathcal{R}$ employs $M$ codebooks in the quantization head $\mathcal{Q}(\mathbf{\hat{x}};\theta_{\mathcal{Q}})$ of $\{C_1,...,C_{M}\}\subset\mathcal{Q}$, where $C_i$ consists of $K$ codewords $\mathbf{c}\in R^{D/M}$ as $C_m=\{\mathbf{c}_{m1},...,\mathbf{c}_{mK}\}$. PQ is conducted in $\mathcal{Q}$ by dividing the deep feature space into the Cartesian product of multiple subspaces. Every codebook of corresponding subspace exhibits several distinctive characteristics representing the image dataset $\mathcal{X}$. Each codeword belonging to the codebook infers a clustered centroid of a divided deep descriptor, which aims to hold a local pattern that frequently occurs. During quantization, similar properties between images are shared by being assigned to the same codeword, whereas distinguishable features have different codewords. As a result, various distance representations for efficient image retrieval are achieved.

\subsection{Self-supervised Training}
\label{subsection3.2}

For better understanding, we briefly describe the training scheme of SPQ in Algorithm \ref{algorithm1}, where $\theta_{\mathcal{F}}$ and $\theta_{\mathcal{Q}}$ denote trainable parameters of feature extractor and quantization head, respectively, and $\gamma$ is a learning rate. In this case, $\theta_{\mathcal{Q}}$ represents a collection of codebooks. The training scheme and quantization process of SPQ are detailed below.

First of all, to conduct deep learning with $\mathcal{F}$ and $\mathcal{Q}$ in an end-to-end manner, and make the whole codewords contribute to training, we need to solve the infeasible derivative calculation of hard assignment quantization. For this, following the method in \cite{PQN}, we introduce soft quantization on the quantization head with the soft quantizer $q_m(\cdot)$ as:

\begin{align}
\mathbf{z}_{nm}=\sum_{k}^{K}\frac{\exp({-\lVert\mathbf{x}_{nm}-\mathbf{c}_{mk}}\rVert^2_2/\tau_q)}{\sum_{k'}^{K}\exp({-\lVert\mathbf{x}_{nm}- \mathbf{c}_{mk'}}\rVert^2_2/\tau_q)}\mathbf{c}_{mk}
\label{equation:Eqn1}
\end{align}

\noindent where $\tau_q$ is a non-negative temperature parameter that scales the input of the softmax, and $\lVert\cdot\rVert^2_2$ denotes the squared Euclidean distance to measure the similarity between inputs. In this fashion, the sub-quantized descriptor $\mathbf{z}_{nm}=q_m(\mathbf{x}_{nm}; \tau_q, C_m)$ can be regarded as an exponential weighted sum of the codewords that belong to $C_m$. Note that the entire codewords in the codebook are utilized to approximate the quantized output, where the closest codeword contributes the most.

Besides, unlike previous deep PQ approaches~\cite{PQN, GPQ}, we exclude intra-normalization \cite{VLAD} which is known to minimize the impact of burst visual features when concatenating sub-quantized descriptors to obtain the whole product quantized descriptor $\mathbf{\hat{z}}$. Since our SPQ is trained without any human supervision, which assists in finding distinct features, we focus on catching dominant visual features rather than balancing the influence of every codebook.

To learn the deep descriptors and the codewords together, we propose \textit{cross quantized contrastive learning} scheme. Inspired by contrastive learning~\cite{Simclr, CURL, SWAV}, we attempt to compare the deep descriptors and the product quantized descriptors of various views (transformed images). As observed in Figure \ref{fig:Figure2}, the deep descriptor and the product quantized descriptor are treated as correlated if the views are originated from the same image, whereas uncorrelated if the views are originated from the different images. Note that, to increase the generalization capacity of the codewords, the correlation between the deep descriptor and the quantized descriptor of itself ($\mathbf{\hat{x}}_{n}$ and $\mathbf{\hat{z}}_{n}$) is ignored. This is because the contribution of other codewords decreases when the agreement between the subvector and the nearest codeword is maximized.

For a given mini-batch of size $N_B$, we randomly sample $N_B$ examples from the database $\mathcal{X}$ and apply a random combination of augmentation techniques to each image twice to generate $2N_B$ data points (views). Inspired from \cite{Collaborative, Simclr, SWAV}, we take into account that two separate views of the same image $(\tilde{x}_i,\tilde{x}_j)$ are correlated, and the other $2(N_B-1)$ views originating from different images within a mini-batch are uncorrelated. On this assumption, we design a cross quantized contrastive loss function to learn the correlated pair of examples $(i,j)$ as:

\begin{equation}
\ell_{(i,j)} = -\log\frac{\exp\left(\mathcal{S}(i,j)/\tau_{cqc}\right)}{\sum_{n=1}^{N_B}\mathds{1}_{[n'\neq j]}\exp\left(\mathcal{S}(i,n')/\tau_{cqc}\right)}
\label{equation:Eqn2}
\end{equation}

\noindent where $n' = \begin{cases}
2n-1 & \text{if } j \text{ is odd} \\
2n & \text{else}
\end{cases}$, $\mathcal{S}(i,j)$ denotes a cosine similarity between $\mathbf{\hat{x}}_i$ and $\mathbf{\hat{z}}_j$, $\tau_{cqc}$ is a non-negative temperature parameter, and $\mathds{1}_{[n'\neq j]}\in{\{0,1\}}$ is an indicator that evaluates to 1 iff $n'\neq j$. Notably, to reduce redundancy between $\mathbf{\hat{x}}_i$ and $\mathbf{\hat{z}}_i$ which are similar to each other, the loss is computed for the half of the uncorrelated samples in the batch. The cosine similarity is used as a distance measure to avoid the norm deviations between $\mathbf{\hat{x}}$ and $\mathbf{\hat{z}}$ .

\begin{algorithm}[!h]
\caption{SPQ's main learning algorithm.}
\begin{algorithmic}[1]
\label{algorithm1}

\REQUIRE Trainable parameters : $\theta_{\mathcal{F}}, \theta_{\mathcal{Q}}$, batch size $N_B$
\FOR{sampled mini-batch $\{x_n\}_{n=1}^{N_B}$}
\FOR{$n$ in \{1,\dots,$N_B$\}}
\STATE draw two transformations $t_{2n-1}\sim\mathcal{T}, t_{2n}\sim\mathcal{T}$
\STATE $\tilde{x}_{2n-1} \gets t_{2n-1}(x_n)$
\STATE $\tilde{x}_{2n} \gets t_{2n}(x_n)$
\STATE $\mathbf{\hat{x}}_{2n-1}, \mathbf{\hat{x}}_{2n}$ = $\mathcal{F}(\tilde{x}_{2n-1}), \mathcal{F}(\tilde{x}_{2n})$
\STATE $\mathbf{\hat{z}}_{2n-1}, \mathbf{\hat{z}}_{2n}$ = $\mathcal{Q}(\mathbf{\hat{x}}_{2n-1}),\mathcal{Q}(\mathbf{\hat{x}}_{2n})$
\ENDFOR
\FOR{$i$ in \{1,\dots,$N_B$\} and $j$ in \{1,\dots,$N_B$\}}
\STATE $\mathcal{S}(2i-1,2j)=\mathbf{\hat{x}}_{2i-1}^{T}\mathbf{\hat{z}}_{2j}/(\lVert\mathbf{\hat{x}}_{2i-1}\rVert\lVert\mathbf{\hat{z}}_{2j}\rVert)$
\STATE $\mathcal{S}(2i,2j-1)=\mathbf{\hat{x}}_{2i}^{T}\mathbf{\hat{z}}_{2j-1}/(\lVert\mathbf{\hat{x}}_{2i}\rVert\lVert\mathbf{\hat{z}}_{2j-1}\rVert)$
\ENDFOR
\STATE $\mathcal{L}_{cqc}=\frac{1}{2N_B}\sum_{n=1}^{N_B}\left(\ell_{(2n-1,2n)} + \ell_{(2n, 2n-1)}\right)$
\STATE $\theta_\mathcal{F} \leftarrow{\theta_\mathcal{F} - \gamma \frac{\partial \mathcal{L}_{cqc}}{\partial \theta_\mathcal{F}}}$
\STATE $\theta_\mathcal{Q} \leftarrow{\theta_\mathcal{Q} - \gamma \frac{\partial \mathcal{L}_{cqc}}{\partial \theta_\mathcal{Q}}}$
\ENDFOR
\ENSURE Updated $\theta_\mathcal{F}, \theta_\mathcal{Q}$

\end{algorithmic}
\end{algorithm}

Concerning the data augmentation for generating various views, we employ five popular techniques: (1) resized crop to treat local, global, and adjacent views, (2) horizontal flip to handle mirrored inputs, (3) color jitter to deal with color distortions, (4) grayscale to focus more on intensity, and (5) Gaussian blur to cope with noise in the image. The default setup is directly taken from \cite{Simclr}, where all transformations are randomly applied in a sequential manner (1-5). Exceptionally, we modify color jitter strength as 0.5 to fit in SPQ, following the empirical observation. In the end, SPQ is capable of interpreting contents in the image by contrasting different views of images in a self-supervised way.

\subsection{Retrieval}
\label{subsection3.3}

Image retrieval is performed in two steps, similar to PQ \cite{PQ}. First, the retrieval database consisting of binary codes is configured with a dataset $\mathcal{X}_{g}=\{x_n\}^{N_g}_{n=1}$ of $N_g$ gallery images. By employing $\mathcal{F}$, the deep descriptor $\mathbf{\hat{x}}_n$ is obtained from $x_n$ and divided into $M$ equal-length subvectors as $\mathbf{\hat{x}}_n=\left[\mathbf{x}_{n1},...,\mathbf{x}_{nM}\right]$. Then, the nearest codeword of each subvector $\mathbf{x}_{nm}$ is searched by computing the squared Euclidean distance ($\lVert\cdot\rVert^2_2$) between the subvector and every codeword in the codebook $C_m$. Then, the index of the nearest codeword $k^{*}$ is formatted as a binary code to generate a sub-binary code $\mathbf{b}_{nm}$. Finally, all the sub-binary codes are concatenated to generate the $M{\cdot}\log_2(K)$-bits binary code $\hat{\mathbf{b}}_n$, where $\hat{\mathbf{b}}_n=[\mathbf{b}_{n1},...,\mathbf{b}_{nM}]$. We repeat this process for all gallery images to build a binary encoded retrieval database.

Moving on to the retrieval stage, we apply the same $\mathcal{F}$ and dividing process on a query image $x_q$ to extract $\mathbf{\hat{x}}_q$ and a set of its subvectors as $\mathbf{\hat{x}}_q=\left[\mathbf{x}_{q1},...,\mathbf{x}_{qM}\right]$. The Euclidean distance is utilized to measure the similarity between the subvectors and every codeword of all codebooks to construct a pre-computed look-up table. The distance calculation between the query and the gallery is asymmetrically approximated and accelerated by summing up the look-up results.

\section{Experiments}
\label{section4}

\subsection{Datasets}
\label{subsection4.1}

To evaluate the performance of SPQ, we conduct comprehensive experiments on three public benchmark datasets, following experimental protocols in recent unsupervised deep image retrieval methods \cite{BinGAN, DistillHash, TBH}.

\noindent \textbf{CIFAR-10} \cite{CIFAR-10} contains 60,000 images with the size of $32 \times 32$ in 10 class labels, and each class has 6,000 images. We select 5,000 images per class as a training set, 100 images per class as a query set. The entire training set of 50,000 images are utilized to build a retrieval database.

\noindent \textbf{FLICKR25K} \cite{FLICKR} consists 25,000 images with various resolutions collected from the Flickr website. Every image is manually annotated with at least one of the 24 semantic labels. We randomly take 2,000 images as a query set and employ the remaining 23,000 images to build a retrieval database, of which 5,000 images are utilized for training.

\noindent \textbf{NUS-WIDE} \cite{NUS-WIDE} has nearly 270,000 images with various resolutions in 81 unique labels, where each image belongs to one or more labels. We pick out images containing the 21 most frequent categories to perform experiments with a total of 169,643. We randomly choose a total of 10,500 images as a training set with each category being at least 500, a total of 2,100 images as a query set with each category being at least 100, and the rest images as a retrieval database. 

\begin{table}[!h]
\caption{Composition of three benchmark datasets.}
\centering
\begin{adjustbox}{width=0.45\textwidth}
\begin{tabular}{c|c|c|c|c}
\toprule 
Dataset & \# Train & \# Query & \# Retrieval & \# Class\\
\midrule
\midrule 
CIFAR-10 & 50,000 & 10,000 & 50,000 & 10\\
FLICKR25K & 5,000 & 2,000 & 23,000  & 24\\
NUS-WIDE & 10,500 & 2,100 & 157,043 & 21\\
\bottomrule
\end{tabular}
\end{adjustbox}

\label{table:Table1}
\end{table}

\begin{table*}[!t]
\caption{mAP scores of different retrieval methods on three benchmark datasets.}
\centering
\begin{adjustbox}{width=0.8\textwidth}
\begin{tabular}{clccccccccc}
\toprule
\multicolumn{2}{c}{\multirow{2}{*}{Method}} & \multicolumn{3}{c}{CIFAR-10}  & \multicolumn{3}{c}{FLICKR25K}  & \multicolumn{3}{c}{NUS-WIDE} \\
\multicolumn{2}{c}{}                        & 16-bits          & 32-bits          & 64-bits          & 16-bits          & 32-bits          & 64-bits  & 16-bits          & 32-bits          & 64-bits \\ \midrule\midrule\multicolumn{11}{c}{\textit{Shallow Methods without Deep Learning}} \\\midrule
\multicolumn{2}{c}{LSH \cite{LSH}}                     & 0.132          & 0.158         & 0.167         & 0.583          & 0.589          & 0.593          & 0.432          & 0.441   &0.443       \\ \midrule
\multicolumn{2}{c}{SH \cite{SH}}                     & 0.272          & 0.285         & 0.300         & 0.591          & 0.592          & 0.602          & 0.510          & 0.512   &0.518       \\ \midrule
\multicolumn{2}{c}{ITQ \cite{ITQ}}                     & 0.305          & 0.325         & 0.349         & 0.610          & 0.622          & 0.624          & 0.627          & 0.645   &0.664       \\ \midrule
\multicolumn{2}{c}{PQ \cite{PQ}}                     & 0.237          & 0.259         & 0.272         & 0.601          & 0.612          & 0.626          & 0.452          & 0.464   &0.479       \\ \midrule
\multicolumn{2}{c}{OPQ \cite{OPQ}}                     & 0.297          & 0.314         & 0.323         & 0.620          & 0.626          & 0.629          & 0.565          & 0.579   &0.598       \\ \midrule
\multicolumn{2}{c}{LOPQ \cite{LOPQ}}                     & 0.314          & 0.320         & 0.355         & 0.614          & 0.634          & 0.635          & 0.620          & 0.655   &0.670       \\ \midrule\midrule\multicolumn{11}{c}{\textit{Deep Semi-unsupervised Methods}} \\\midrule
\multicolumn{2}{c}{DeepBit \cite{DeepBit}}                     & 0.220          & 0.249         & 0.277         & 0.593          & 0.593          & 0.620          & 0.454          & 0.463   &0.477       \\ \midrule
\multicolumn{2}{c}{GreedyHash \cite{GreedyHash}}                     & 0.448          & 0.473         & 0.501         & 0.689          & 0.699          & 0.701          & 0.633          & 0.691   &0.731       \\ \midrule
\multicolumn{2}{c}{DVB \cite{DVB}}                     & 0.403          & 0.422         & 0.446         & 0.614          & 0.655          & 0.658          & 0.677          & 0.632   &0.665       \\ \midrule
\multicolumn{2}{c}{DistillHash \cite{DistillHash}}                     & 0.454          & 0.469         & 0.489         & 0.696          & 0.706          & 0.708          & 0.667          & 0.675   &0.677       \\ \midrule
\multicolumn{2}{c}{TBH \cite{TBH}}                     & 0.532          & 0.573         & 0.578         & 0.702          & 0.714          & 0.720          & 0.717          & 0.725   &0.735       \\ \midrule\midrule\multicolumn{11}{c}{\textit{Deep truly unsupervised Methods}} \\\midrule
\multicolumn{2}{c}{SGH \cite{SGH}}                     & 0.435          & 0.437         & 0.433         & 0.616          & 0.628          & 0.625          & 0.593          & 0.590   &0.607       \\ \midrule
\multicolumn{2}{c}{HashGAN \cite{HashGAN}}                     & 0.447          & 0.463         & 0.481         & -          & -          & -          & -          & -   & -       \\ \midrule
\multicolumn{2}{c}{BinGAN \cite{BinGAN}}                     & 0.476          & 0.512         & 0.520         & 0.663          & 0.679          & 0.688          & 0.654          & 0.709   &0.713       \\ \midrule
\multicolumn{2}{c}{BGAN \cite{BGAN}}                     & 0.525          & 0.531         & 0.562         & 0.671          & 0.686          & 0.695          & 0.684          & 0.714   &0.730       \\ \midrule
\multicolumn{2}{c}{\textbf{SPQ} (Ours)}                    & \textbf{0.768} & \textbf{0.793} & \textbf{0.812}   & \textbf{0.757} & \textbf{0.769} & \textbf{0.778} &  \textbf{0.766} &  \textbf{0.774}   & \textbf{0.785}   \\ \bottomrule
\end{tabular}
\end{adjustbox}

\label{table:Table2}
\end{table*}

\subsection{Experimental Settings}
\label{subsection4.2}

\noindent \textbf{Evaluation Metrics.} We employ mean Average Precision (mAP) to evaluate the retrieval performance. Specifically, in the case of multi-label image retrieval on FLICKR25K and NUS-WIDE dataset, it is considered relevant even if only one of the labels matches. We vary the number of bits allocated to the binary code as $\{16, 32, 64\}$ to measure the mAP scores of the retrieval approaches, mAP@1,000 for CIFAR-10 dataset and mAP@5,000 for FLICKR25K and NUS-WIDE datasets, following the evaluation method in \cite{TBH, DistillHash}. In addition, by employing 64-bits hash codes of different algorithms, we draw Precision-Recall curves (PR) to compare the precisions at different recall levels and report Precision curves with respect to 1,000 top returned samples (P@1,000) to contrast the ratio of results retrieved correctly.

\noindent \textbf{Implementation details.} There are three baseline approaches that we categorize to make a comparison. Specifically, (1) shallow methods without deep learning, based on hashing: LSH \cite{LSH}, SH \cite{SH}, ITQ \cite{ITQ}, and based on product quantization: PQ \cite{PQ}, OPQ \cite{OPQ} LOPQ \cite{LOPQ}, (2) deep semi-unsupervised methods: DeepBit \cite{DeepBit}, GreedyHash \cite{GreedyHash}, DVB \cite{DVB}, DistillHash \cite{DistillHash}, TBH \cite{TBH}, and (3) deep truly unsupervised methods: SGH \cite{SGH}, HashGAN \cite{HashGAN}, BinGAN \cite{BinGAN}, BGAN \cite{BGAN}. The terms ``semi'' and ``truly'' indicate whether the pretrained model weights are utilized or not. Both semi and truly training conditions can be applied to SPQ; however, we take the truly unsupervised model that has the advantage of not requiring human supervision as the baseline.

To evaluate the shallow and deep semi-unsupervised methods, we employ ImageNet pretrained model weights of AlexNet \cite{AlexNet} or VGG16 \cite{VGG} to utilize $fc\_7$ features, following the experimental settings of \cite{DistillHash, DVB, TBH}. Since those models take only fixed-size inputs, we need to resize all images to $224 \times 224$, by upscaling the small images and downsampling the large ones. In the case of evaluating deep truly unsupervised methods, including SPQ, the same resized images of FLICKR25K and NUS-WIDE datasets are used for simplicity, and the original resolution images of CIFAR-10 are used to reduce computational load.

Our implementation of SPQ is based on PyTorch with NVIDIA Tesla V100 32GB Tensor Core GPU. Following the observations in recent self-supervised learning studies \cite{Simclr, SWAV}, we set the baseline network architecture as a standard ResNet50 \cite{ResNet} for FLICKR25K and NUS-WIDE datasets. In the case of the CIFAR-10 dataset with much smaller images, we set the baseline as a standard ResNet18 \cite{ResNet}, and modify the number of filters as same as ResNet50.

For network training, we adopt Adam \cite{Adam} and decay the learning rate with the cosine scheduling without restarts \cite{CosLR} and set the batch size $N_B$ as 256. We fix the dimension of the subvector $\mathrm{x}$ and the codeword $\mathrm{c}$ to $D/M=16$, and also the number of codewords to $K=2^4$. Consequently, the number codebooks $M$ is changed to $\{4, 8, 16\}$ because $M{\cdot}\log_2(K)$-bits are needed to obtain $\{16, 32, 64\}$-bits binary code. The temperature parameter $\tau_{q}$ and $\tau_{cqc}$ are set as 0.2 and 0.5, respectively. Data augmentation is operated with Kornia \cite{KORNIA} library, and each transformation is applied with the same probability as the settings in \cite{Simclr}. 
\begin{figure*}[!t]
\centering
\subfigure[CIFAR-10]{
\includegraphics[width=0.31\linewidth]{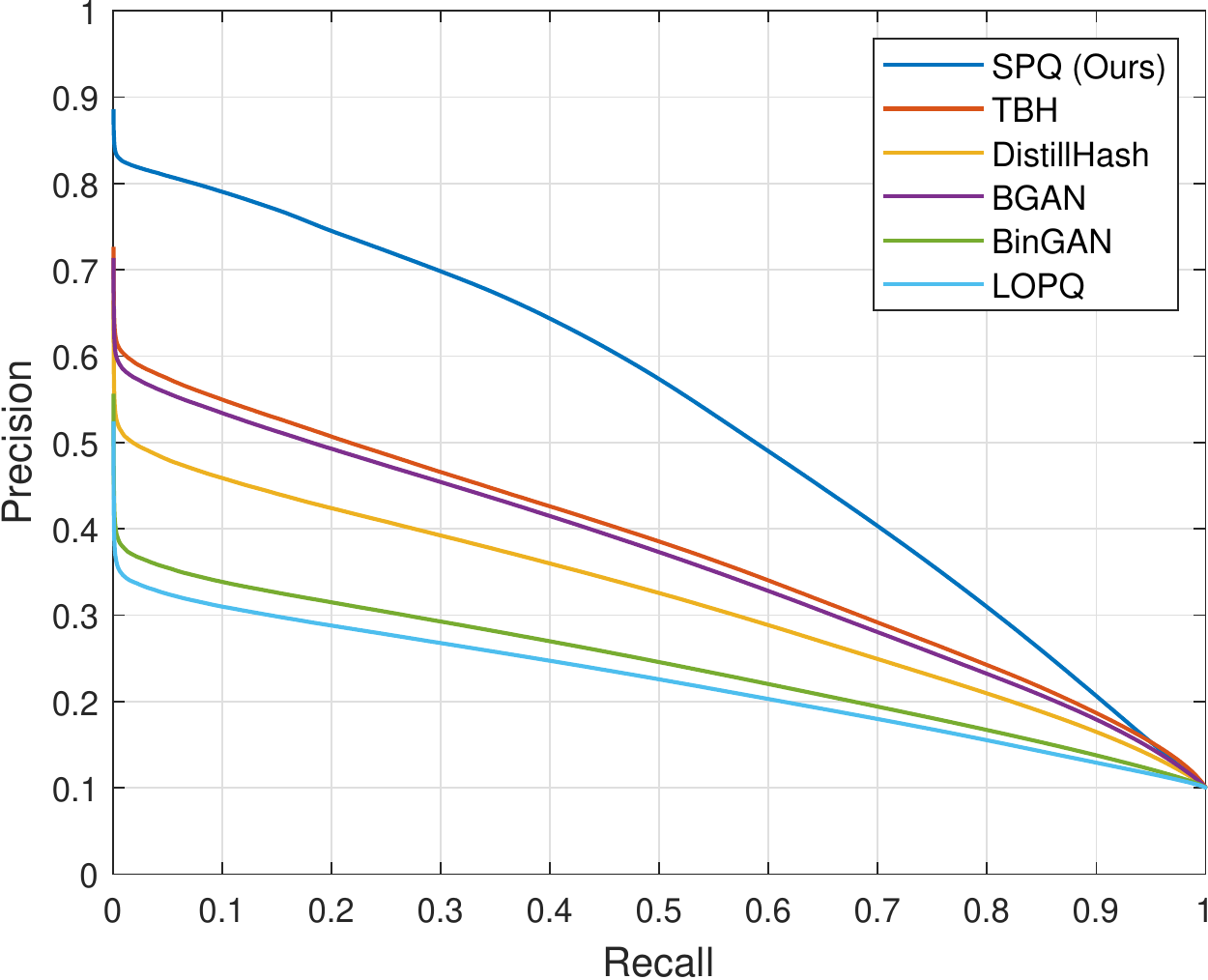}
}
\subfigure[FLICKR25K]{
\includegraphics[width=0.31\linewidth]{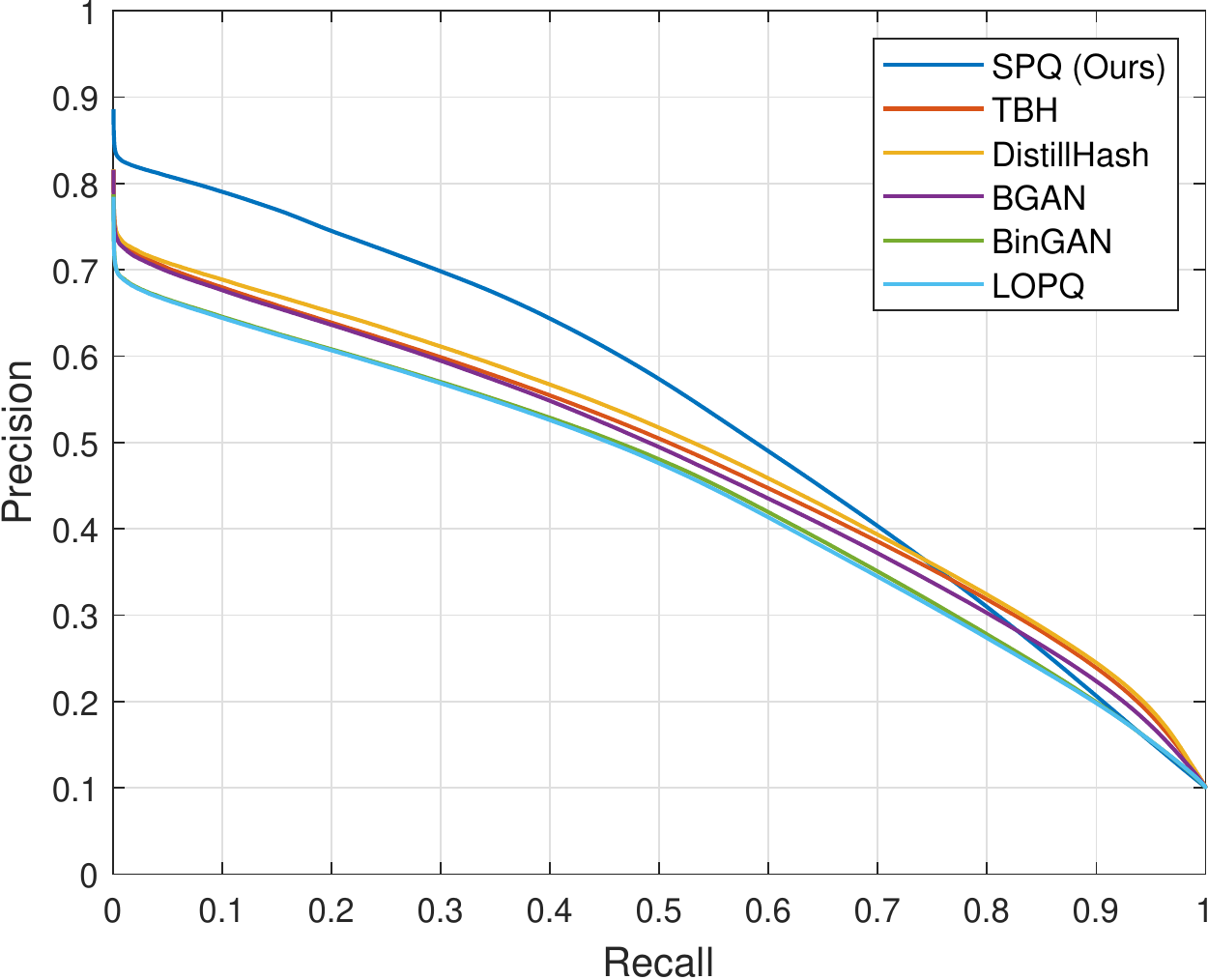}
}
\subfigure[NUS-WIDE]{
\includegraphics[width=0.31\linewidth]{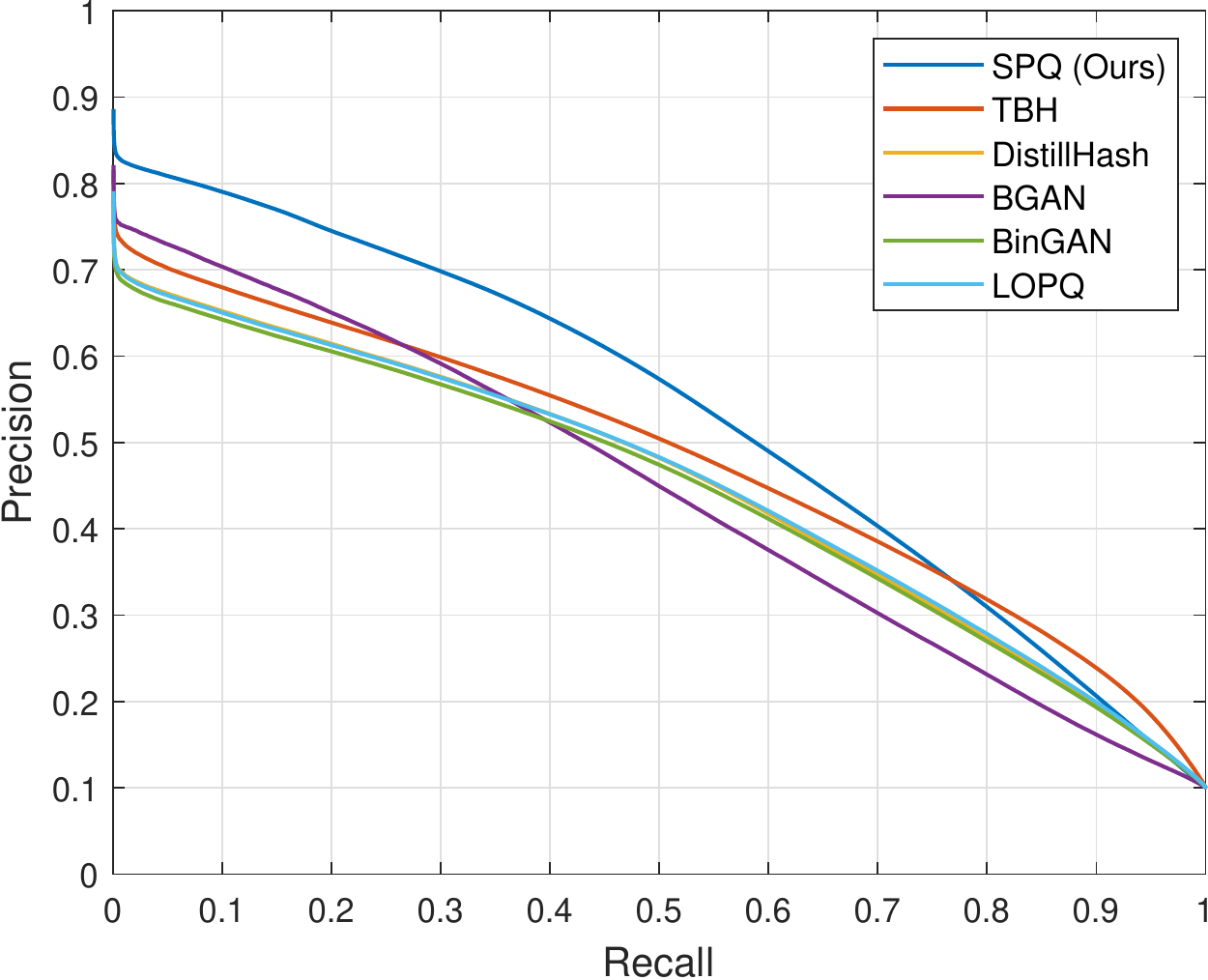}
}
\caption{Precision-Recall curves on three benchmark datasets with binary codes @ 64-bits.} 
\label{fig:Figure3}
\end{figure*}

\begin{figure*}[!t]
\centering
\subfigure[CIFAR-10]{
\includegraphics[width=0.31\linewidth]{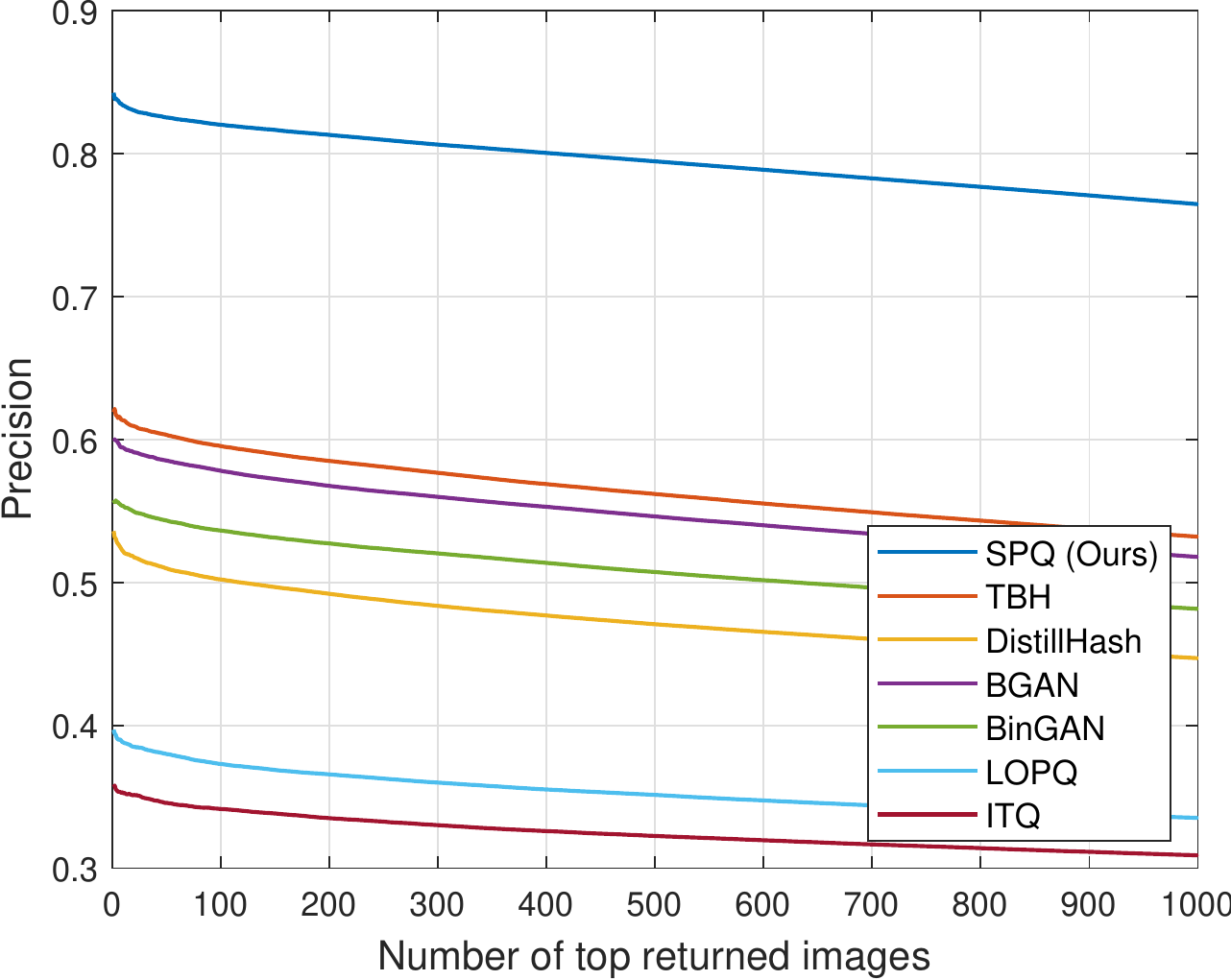}
}
\subfigure[FLICKR25K]{
\includegraphics[width=0.31\linewidth]{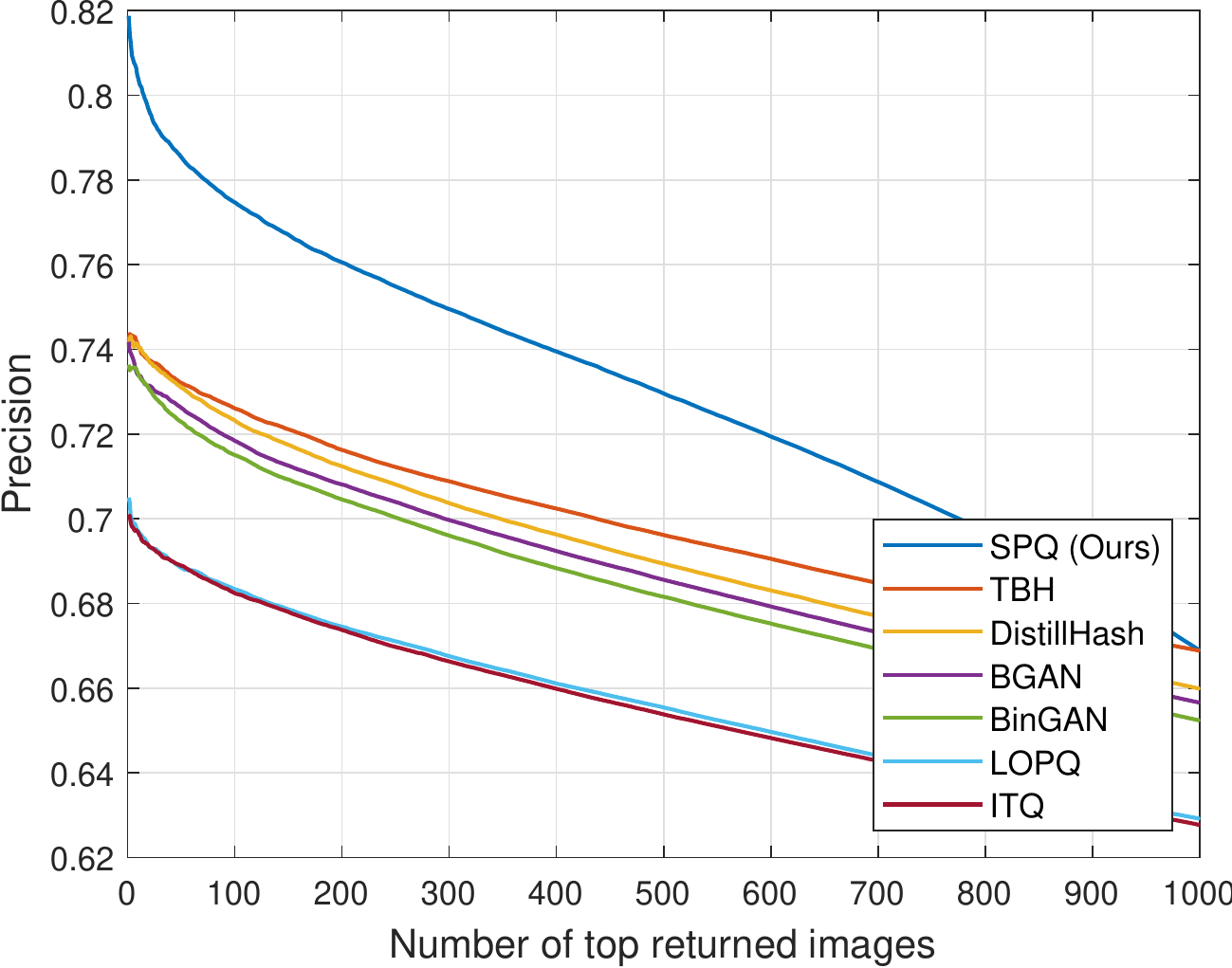}
}
\subfigure[NUS-WIDE]{
\includegraphics[width=0.31\linewidth]{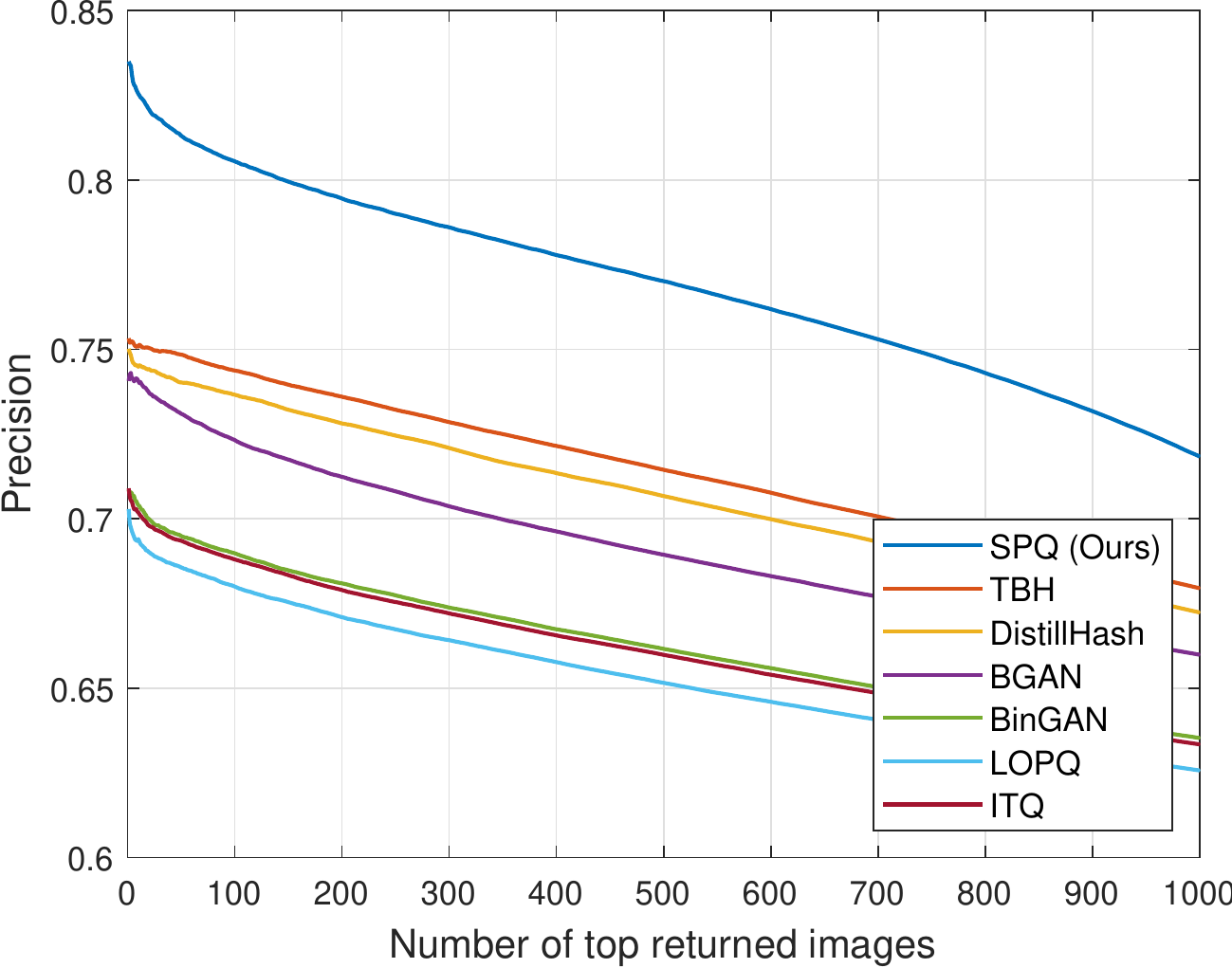}
}
\caption{Precision@top-1000 curves on three benchmark datasets with binary codes @ 64-bits.} 
\label{fig:Figure4}
\end{figure*}

\subsection{Results}
\label{subsection4.3}

The mAP results on three different image retrieval datasets are listed in Table \ref{table:Table2}, showing that SPQ substantially outperforms all the compared methods in every bit-length. Additionally, referring to Figures \ref{fig:Figure3} and \ref{fig:Figure4}, SPQ is demonstrated to be the most desirable retrieval system.

First of all, compared to the best shallow method LOPQ \cite{LOPQ}, SPQ reveals a performance improvement of more than 46\%p, 13\%p, and 11.6\%p in the average mAP on CIFAR-10, FLICKR25K, and NUS-WIDE, respectively. The reason for the more pronounced difference for CIFAR-10 is because the shallow methods involve an unnecessary upscaling process to utilize the ImageNet pretrained deep features. SPQ has an advantage over shallow methods in that various and suitable neural architectures can be accommodated for feature extraction and end-to-end learning.

Second, in contrast to the best deep semi-unsupervised method TBH \cite{TBH}, SPQ yields 23\%p, 4.6\%p, and 3.9\%p higher average mAP scores on CIFAR-10, FLICKR25K, and NUS-WIDE, respectively. Even in the absence of prior information such as pretrained model weights, SPQ well distinguishes the contents within the image by comparing multiple views of training samples.

Lastly, even with the truly unsupervised setup, SPQ achieves state-of-the-art retrieval accuracy. Specifically, unlike previous hashing-based truly unsupervised methods, SPQ introduces differentiable product quantization to the unsupervised image retrieval system for the first time. By considering cross-similarity between different views in a self-supervised way, deep descriptors and codewords are allowed to be discriminative.

\begin{figure*}[!t]
\centering
\subfigure[BinGAN \cite{BinGAN}]{
\includegraphics[width=0.32\linewidth]{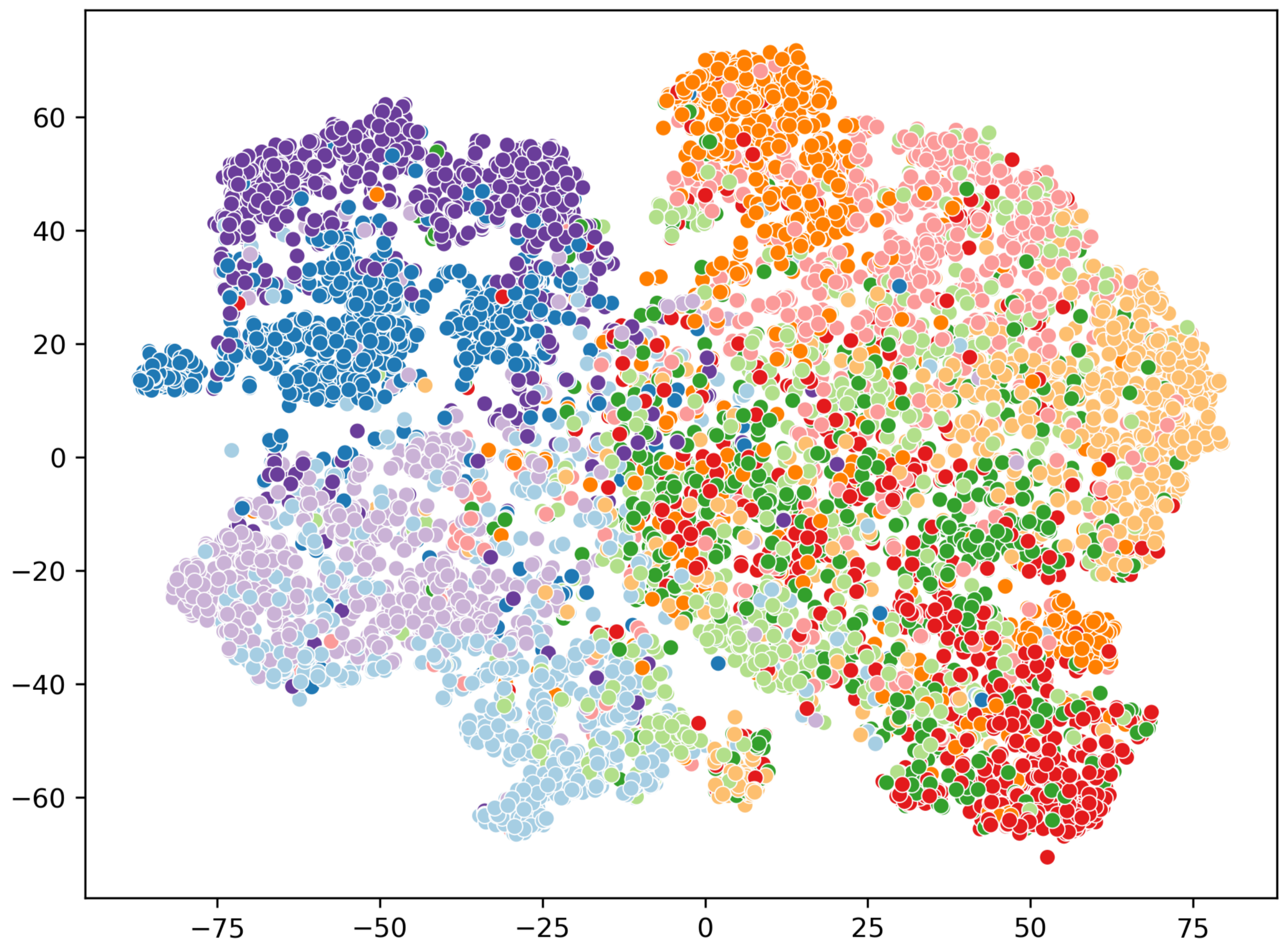}
}
\subfigure[TBH \cite{TBH}]{
\includegraphics[width=0.32\linewidth]{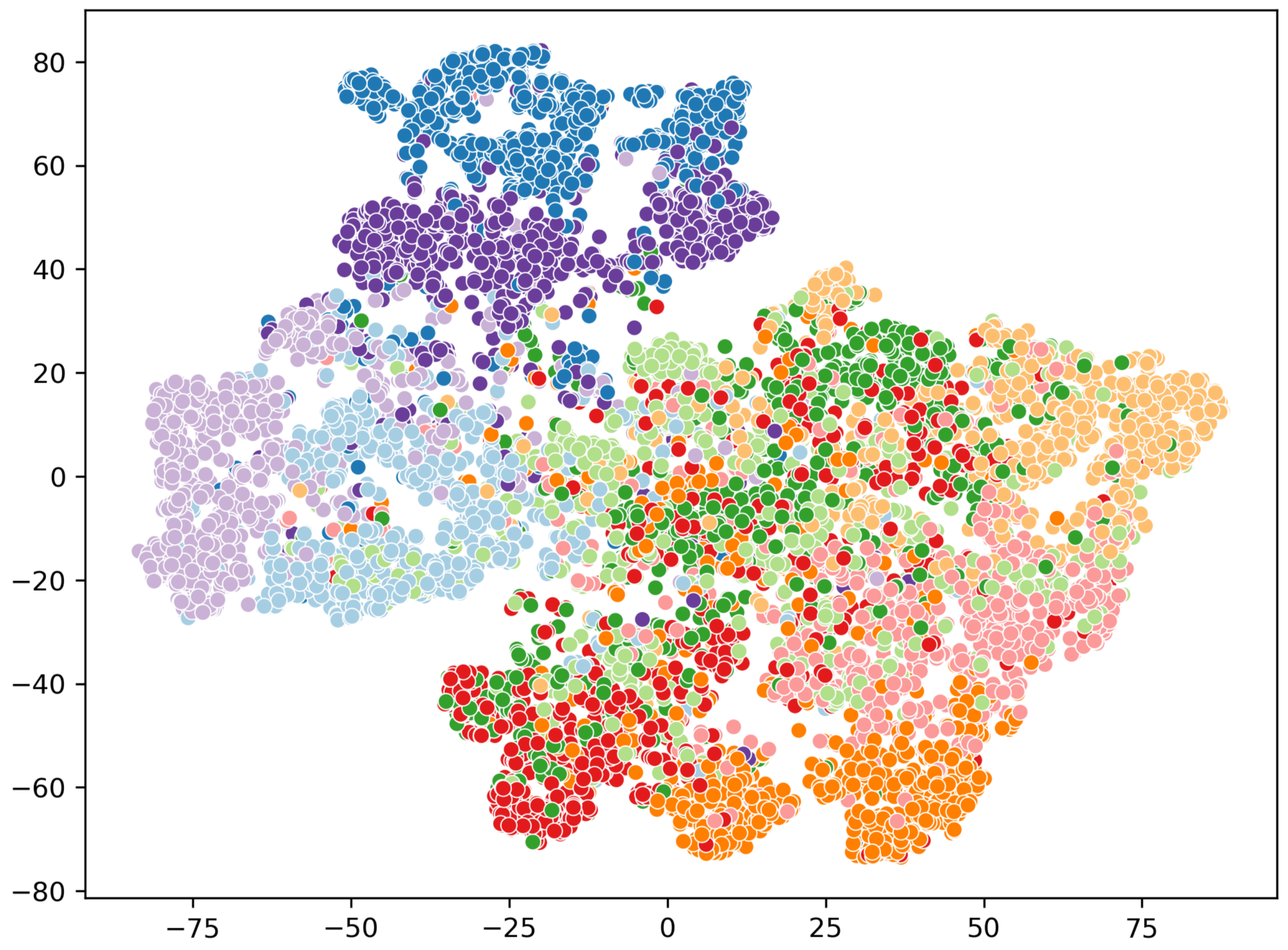}
}
\subfigure[SPQ (Ours)]{
\includegraphics[width=0.32\linewidth]{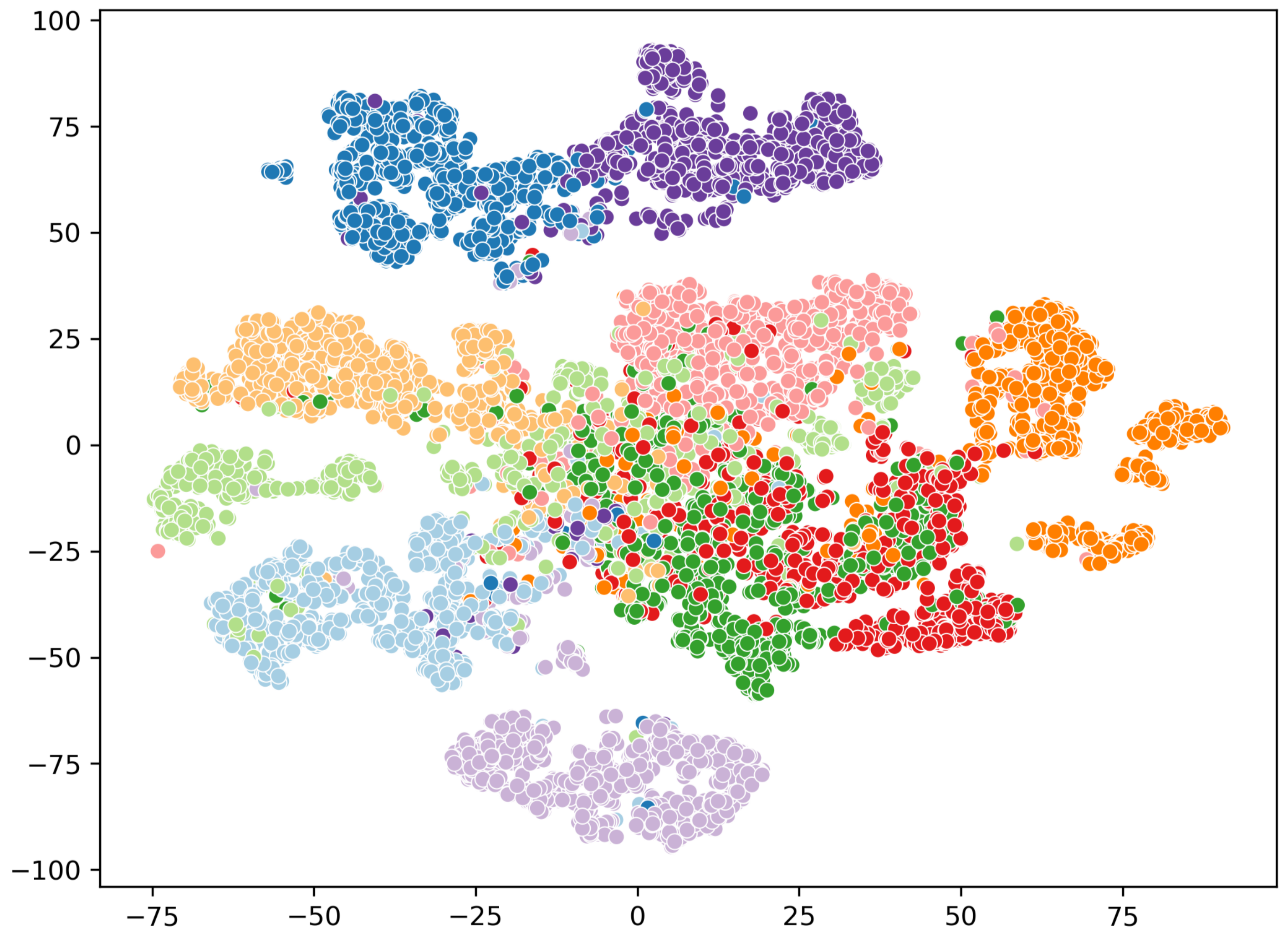}
}
\caption{t-SNE visualization of deep representations learned by BinGAN, TBH, and SPQ on CIFAR-10 query set respectively.} 
\label{fig:Figure5}
\end{figure*}

\subsection{Empirical Analysis}
\subsubsection{Ablation Study}
We configure five variants of SPQ to investigate: (1) \textit{SPQ-C} that replaces cross-quantized contrastive learning with contrastive learning by comparing $\mathbf{\hat{z}}_i$ and $\mathbf{\hat{z}}_j$, (2) \textit{SPQ-H}, employing hard quantization instead of soft quantization, (3) \textit{SPQ-Q} that employs standard vector quantization, which does not divide the feature space and directly utilize entire feature vector to build the codebook, (4) \textit{SPQ-S} that exploits pretrained model weights to conduct deep semi-unsupervised image retrieval, and (5)\textit{SPQ-V}, utilizing VGG16 network architecture as the baseline.

\begin{table}[!t]
\caption{mAP scores of the previous best method, SPQ and its variants on three benchmark datasets @ 32-bits.}
\centering
\begin{adjustbox}{width=0.42\textwidth}
\begin{tabular}{cccc}
\toprule 
Method & CIFAR-10 & FLICKR25K & NUS-WIDE \\ 
\midrule
\midrule
TBH \cite{TBH} & 0.573  &0.714  &0.725\\ \midrule
SPQ-C & 0.763  &0.751  &0.756\\ \midrule
SPQ-H & 0.745  &0.736  &0.742\\ \midrule
SPQ-Q & 0.734  &0.733  &0.738\\ \midrule
SPQ-S & \textbf{0.814}  &\textbf{0.781}  &\textbf{0.788}\\ \midrule
SPQ-V & 0.761  &0.749  &0.753\\ \midrule
SPQ & 0.793  & 0.769  & 0.774\\
\bottomrule
\end{tabular}
\end{adjustbox}
\label{table:Table3}
\end{table}

As reported in Table \ref{table:Table3}, we can observe that each component of SPQ contributes sufficiently to performance improvement. Comparison with SPQ-C confirms that considering cross-similarity rather than comparing quantized outputs delivers more efficient image retrieval results. From the results of SPQ-H, we find that soft quantization is more suitable for learning codewords. The retrieval outcomes with SPQ-Q, which shows the biggest performance gap with SPQ, explain that product quantization leads to accomplishing precise search results by increasing the amount of distance representation. Notably, SPQ-S, which utilizes ImageNet pretrained model weights for network initialization, outperforms truly unsupervised SPQ. In this observation, we can see that although SPQ demonstrates the best retrieval accuracy without any human guidance, better results can be obtained with some label information. Although SPQ-V is inferior to ResNet-based SPQ, its performance still surpasses existing state-of-the-art retrieval algorithms, which proves the excellence of the PQ-based self-supervised learning scheme.

\begin{figure}[!t]
\centering
\subfigure[$\tau_q$ vs. color jitter strength.]{
\includegraphics[width=0.46\linewidth]{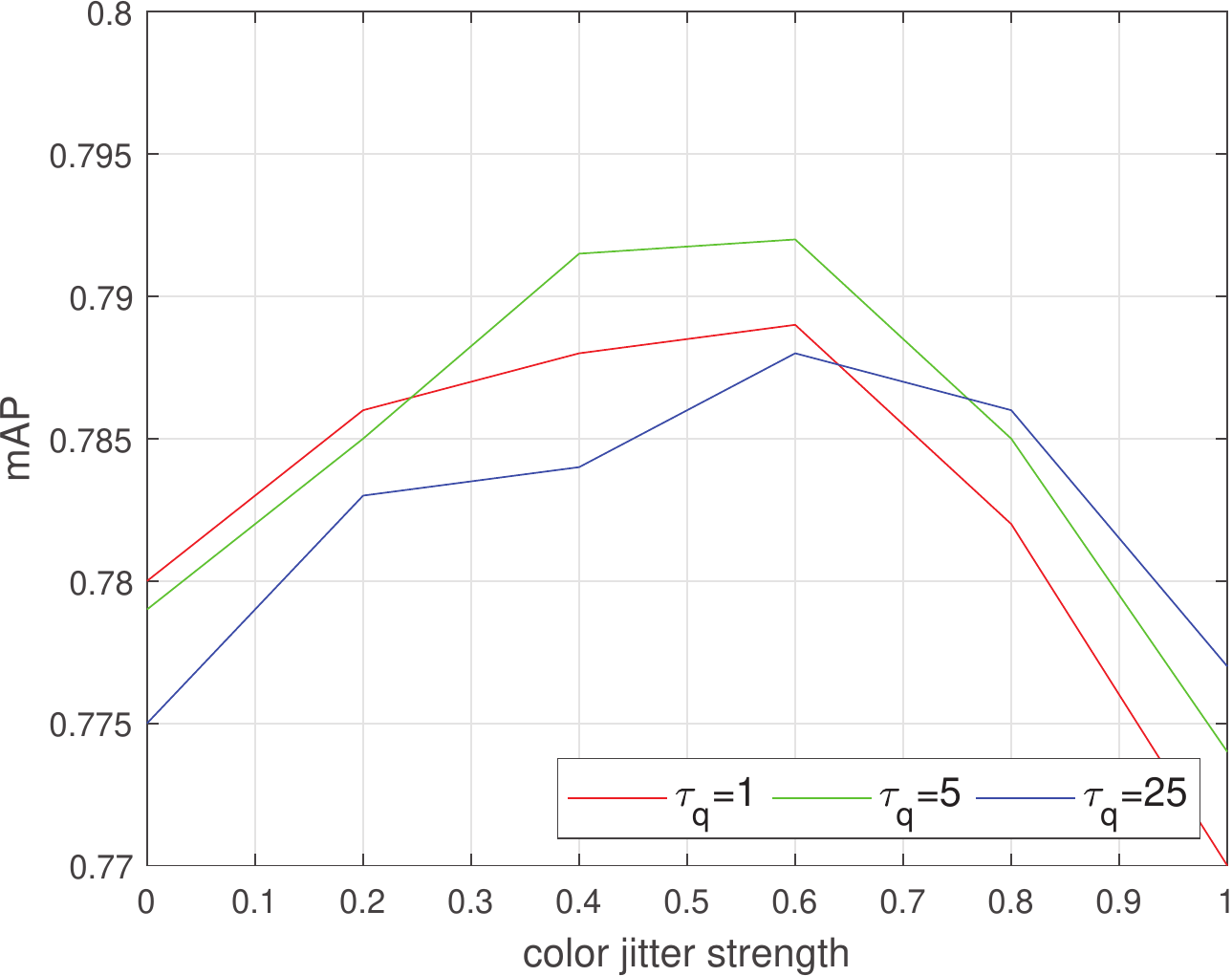}
}
\subfigure[$\tau_{cqc}$ vs. color jitter strength.]{
\includegraphics[width=0.46\linewidth]{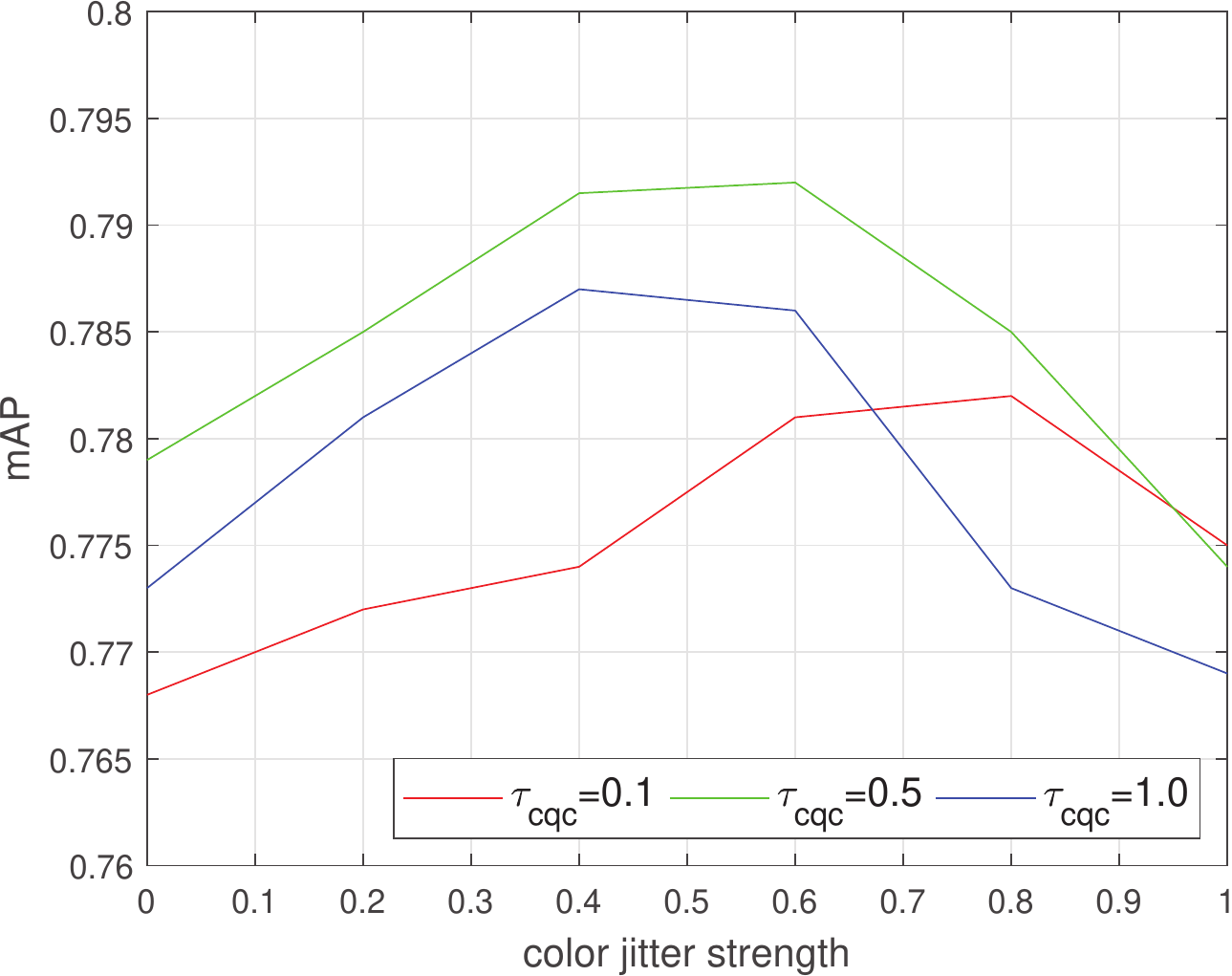}
}
\caption{Sensitivity investigation of hyper-parameters according to the color jitter strengths on CIFAR-10 @ 32-bits.} 
\label{fig:Figure6}
\end{figure}

\begin{figure}[!t]
\centering
\includegraphics[width=0.99\linewidth]{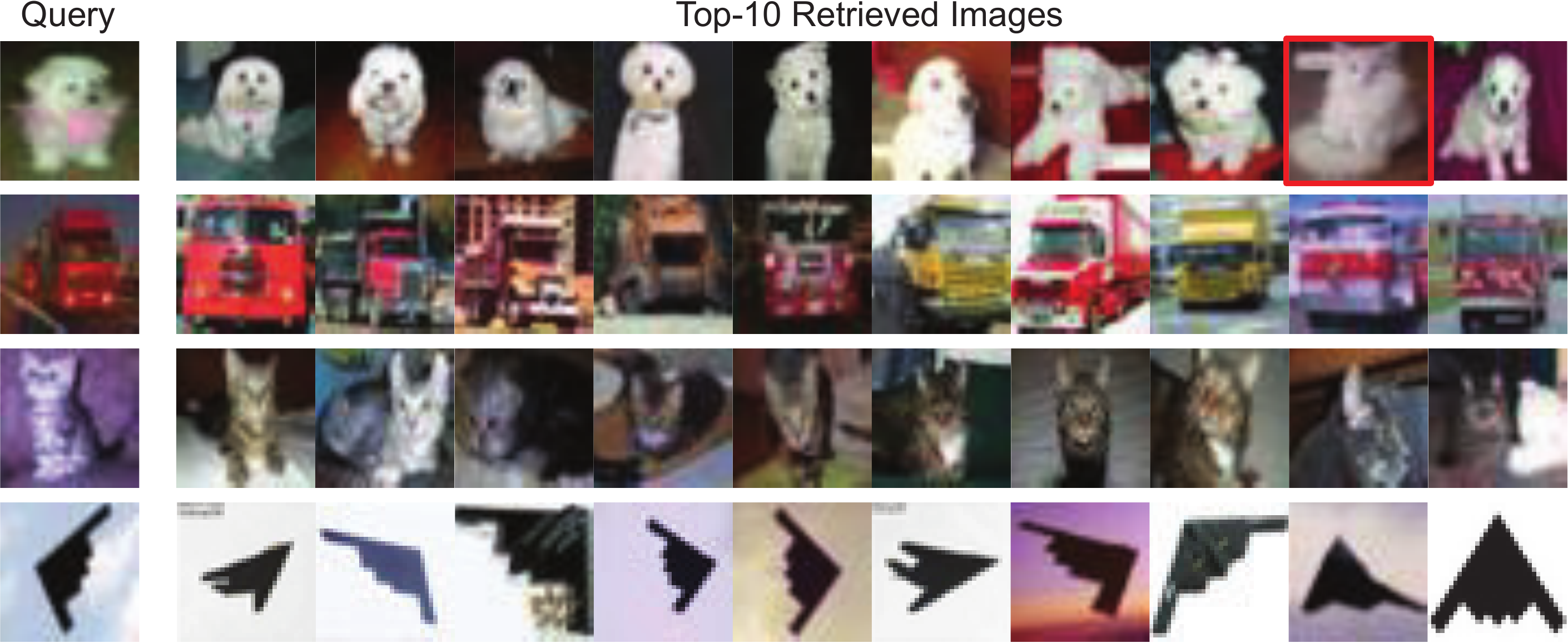}
\caption{SPQ retrieval results on CIFAR-10 @ 32-bits.}
\label{fig:Figure7}
\end{figure}

Besides, we explore an hyper-parameter ($\tau_q$ and $\tau_{cqc}$) sensitivity according to the color jitter strength in Figure \ref{fig:Figure6}. In general, the difference in performance due to the change of hyper-parameters is insignificant; however, the effect of the color jitter strength is pronounced. As a result, we confirm that SPQ is robust to hyper-parameters, and input data preparation is an important factor.

\subsubsection{Visualization}
As illustrated in Figure \ref{fig:Figure5}, we employ t-SNE \cite{t-SNE} to examine the distribution of deep representations of BinGAN, TBH, and our SPQ, where BinGAN and SPQ are trained under the truly unsupervised setting. Nonetheless, our SPQ scatters data samples most distinctly where each color denotes a different class label. Furthermore, we show the actual returned images in Figure \ref{fig:Figure7}. Interestingly, not only images of the same category but also images with visually similar contents are retrieved, like a cat appears in the dog retrieval results.

\section{Conclusion}

In this paper, we have proposed a novel deep self-supervised learning-based fast image retrieval method, Self-supervised Product Quantization (SPQ) network. By employing a product quantization scheme, we built the first end-to-end unsupervised learning framework for image retrieval. We introduced a cross quantized contrastive learning strategy to learn the deep representations and codewords to discriminate the image contents while clustering local patterns at the same time. Despite the absence of any supervised label information, our SPQ yields state-of-the-art retrieval results on three large-scale benchmark datasets. As future research, we expect performance gain by contrasting more views within a batch, which needs a better computing environment. Our code is publicly available at \url{https://github.com/youngkyunJang/SPQ}.

\section{Acknowledgement}

This work was supported in part by the National Research Foundation of Korea (NRF) grant funded by the Korea government (MSIT) (2021R1A2C2007220) and in part by IITP grant funded by the Korea government [No. 2021-0-01343, Artificial Intelligence Graduate School Program (Seoul National University)].

\newpage
{\small
\bibliographystyle{ieee_fullname}
\bibliography{egbib}
}

\end{document}